\documentclass{article} 
\usepackage[preprint]{colm2026_conference}
\usepackage{microtype}
\usepackage{hyperref}
\usepackage{url}
\usepackage{booktabs}
\usepackage{amssymb}

\usepackage{pgfplots}
\usepgfplotslibrary{groupplots}
\pgfplotsset{compat=1.18}

\usepackage{lineno}
\usepackage{tikz}
\definecolor{darkblue}{rgb}{0, 0, 0.5}
\hypersetup{colorlinks=true, citecolor=darkblue, linkcolor=darkblue, urlcolor=darkblue}

\usepackage[most]{tcolorbox}
\usepackage{xcolor}
\usepackage{enumitem}
\usepackage{array}
\definecolor{qstripe}{RGB}{0,90,0} 
\definecolor{qbg}{RGB}{245,245,245} 

\usepackage[table]{xcolor}
\usepackage{booktabs}
\usepackage{array}
\usepackage{pgf}

\newcommand{\scorecell}[1]{%
  \pgfmathparse{int(round(100*#1))}%
  \edef\pct{\pgfmathresult}%
  \cellcolor{red!\the\numexpr100-\pct\relax!green!\pct}#1%
}

\tcbset{
  qcard/.style={
    enhanced,
    colback=qbg,
    colframe=white,
    boxrule=0pt,
    arc=4pt,
    left=8pt,right=8pt,top=6pt,bottom=6pt,
    borderline west={2.2pt}{0pt}{qstripe},
  }
}

\tcbset{
  modelout/.style={
    enhanced, colback=gray!8, colframe=gray!25,
    boxrule=0.4pt, arc=3pt,
    left=6pt, right=6pt, top=4pt, bottom=4pt,
    fontupper=\small
  }
}

\title{Grading the Unspoken: Evaluating Tacit Reasoning in Quantum Field Theory and String Theory with LLMs}

\author{Xingyang Yu\\
Department of physics\\
Virginia Tech\\
\texttt{xingyangy@vt.edu} \\
\And
Yinghuan Zhang \\
Independent Researcher\\
\texttt{yinghuan.flash@gmail.com} \\
\And
Yufei Zhang \\
Independent Researcher\\
\texttt{yufeizhang96@outlook.com}
\And
Zijun Cui \\
Department of Computer Science and Engineering\\
Michigan State University\\
\texttt{cuizijun@msu.edu}
}

\begin{document}

\ifcolmsubmission
\linenumbers
\fi

\maketitle

\begin{abstract}
Large language models have demonstrated impressive performance across many domains of mathematics and physics. 
One natural question is whether such models can support research in highly abstract theoretical fields such as quantum field theory and string theory. Evaluating this possibility faces an immediate challenge: correctness in these domains is layered, tacit, and fundamentally non-binary. Standard answer-matching metrics fail to 
capture whether intermediate conceptual steps are properly reconstructed 
or whether implicit structural constraints are respected.
We construct a compact expert-curated dataset of twelve questions spanning core areas of quantum field theory and string theory, 
and introduce a five-level grading rubric separating statement correctness, key concept awareness, reasoning chain presence, 
tacit step reconstruction, and enrichment. Evaluating multiple contemporary LLMs, we observe near-ceiling performance on explicit derivations within stable conceptual frames, 
but systematic degradation when tasks require reconstruction of omitted reasoning steps 
or reorganization of representations under global consistency constraints. These failures are driven not only
by missing intermediate steps, but by an instability in
representation selection: models often fail to identify the correct
conceptual framing required to resolve implicit tensions. 
We argue that highly abstract theoretical physics provides a uniquely sensitive lens on the epistemic limits of current evaluation paradigms. 
\end{abstract}

\section{Introduction}

In highly abstract areas of theoretical physics, a substantial portion of working knowledge is not acquired through formal textbooks or written derivations alone, but is transmitted implicitly through research practice, informal discussions, and accumulated intuition. 
This phenomenon is reflected throughout the literature (see, e.g.,~\cite{Seiberg:1994pq, polchinski1998string1, Seiberg:1999vs}), where many intermediate steps in standard arguments are routinely omitted, either because they are considered self-evident to experts or because they resist clean formalization.


This form of understanding is commonly described as \emph{tacit knowledge}: knowledge that guides reasoning but is rarely made explicit. 
If AI systems are to assist research in such domains, they must be able not only to reproduce known results, but also to reconstruct the implicit reasoning structures that underlie them.

This raises a central evaluation challenge: How should we evaluate whether AI systems can reconstruct tacit reasoning? This question is fundamentally an evaluation problem. 
Because such reasoning is not explicitly stated, it cannot be captured by standard benchmarks based on final-answer correctness or short derivations.


Large language models (LLMs) have demonstrated strong performance across mathematics, physics, and formal reasoning benchmarks~\citep{zhang2025bayesian,wei2022chain-of-thought}. 
However, these benchmarks primarily test explicit reasoning or result reproduction, 
and therefore provide limited insight into whether models can recover tacit reasoning structures. 
This gap motivates a more targeted evaluation: 
rather than asking whether a model arrives at the correct answer, 
we ask how it reconstructs the reasoning process when key steps are suppressed.

We construct a small expert-curated dataset of questions in quantum field theory and string theory, 
each designed to probe reasoning that is typically implicit in the literature.
We introduce a five-level grading rubric that decomposes correctness into progressively deeper dimensions, 
ranging from statement accuracy to tacit-step reconstruction and conceptual enrichment.

\noindent\textbf{Why quantum field theory and string theory?}

Quantum field theory and string theory provide a particularly suitable testbed for evaluating tacit reasoning for three reasons, each of which amplifies the role of implicit structure in expert reasoning.

\textit{(1) Structural complexity.}
Reasoning in quantum field theory and string theory involves multiple intertwined layers of mathematical structures and theoretical frameworks.
Arguments often require moving between representations while maintaining consistency, with many intermediate steps left implicit (\cite{Maldacena:1997re}).


\textit{(2) Methodological characteristics.}
Expert practice frequently omits technically straightforward but conceptually necessary steps. See~\cite{Vafa:1996xn} for a representative example where key arguments are presented in a highly compressed form.
Reconstructing these steps requires understanding both local manipulations and their role within a broader reasoning chain.


\textit{(3) Epistemic setting.}
These fields are weakly constrained by experiment, placing emphasis on internal consistency and conceptual coherence (\cite{Deligne:1999qp}).
This makes them a clean setting for evaluating reasoning without confounding data-driven factors.


\noindent\textbf{Evaluation challenges.}
These characteristics make quantum field theory and string theory particularly suitable domains for probing LLM reasoning. At the same time, they sharpen a fundamental difficulty: 
even when final answers are correct, it is often unclear whether the underlying reasoning process is valid.
A response may state the correct conclusion while omitting essential intermediate logic, or reproduce formal manipulations without respecting global consistency conditions. 
Standard answer-matching metrics are insufficient to capture these distinctions.
This observation motivated the present study. 
Rather than attempting to benchmark quantum field theory and string theory competence exhaustively, we conduct an early empirical probe designed to examine how contemporary LLMs behave when confronted with tacit, structurally layered reasoning tasks.

\noindent\textbf{Contributions.}
We construct a small expert-curated dataset of twelve questions spanning quantum field theory and string theory, targeting statements whose intermediate reasoning is typically compressed in the literature. 
We introduce a five-level grading rubric that decomposes correctness into progressively deeper dimensions, from statement accuracy to tacit-step reconstruction and conceptual boundary awareness. 
We further analyze model behavior within a two-dimensional reasoning framework that distinguishes mechanism-driven from consistency-driven inference and single-structure from multi-structure reasoning.

\section{Prototypical dataset for tacit knowledge}

We construct a small expert-curated set of twelve questions drawn from quantum field theory and string theory. The goal is not breadth, but diagnostic density: each question targets a commonly cited statement whose intermediate reasoning is typically compressed or omitted in the literature. The dataset spans multiple subfields and reasoning modes, including field theory foundations, symmetry and topological structure,  conformal field theories, supersymmetry, string dualities and  D-brane physics. The detailed list of questions is provided in Appendix~\ref{app:dataset}.

\textbf{Landscape of quantum field theory and string theory.} Table~\ref{tab:subfield_taxonomy} organizes the twelve questions into 
broad subfields; categories overlap intentionally as many questions 
span multiple themes.

\begin{table}[h]
\centering
\begin{tabular}{ll}
\toprule
Broad research area & Questions \\
\midrule
Field theory foundations & Q1, Q4, Q11, Q12 \\
Symmetry and topological structures & Q3, Q4, Q6, Q10, Q11 \\
Conformal and lower-dimensional QFTs & Q2, Q3, Q8, Q9 \\
SUSY and higher dimensional QFTs & Q5, Q6, Q7, Q8 \\
String dualities and D-brane physics & Q7, Q9, Q10 \\
\bottomrule
\end{tabular}
\caption{
Topical coverage of the dataset. Questions may appear in 
multiple categories.
}
\label{tab:subfield_taxonomy}
\end{table}

\textbf{Geometry of reasoning regimes.}
We organize the 12 problems within a two-dimensional reasoning space as 
an interpretive framework for understanding systematic variations in model 
behavior. The first axis, \textbf{mode of inference}, distinguishes 
mechanism-driven reasoning (conclusions follow from expanding explicit 
local derivations) from consistency-driven reasoning (conclusions follow 
from identifying global constraints that restrict admissible results). 
The second axis, \textbf{conceptual organization}, distinguishes reasoning within single frame (the task remains within a fixed conceptual 
frame) from cross-frame reasoning (the task requires restructuring 
the representational framing before any derivation proceeds).

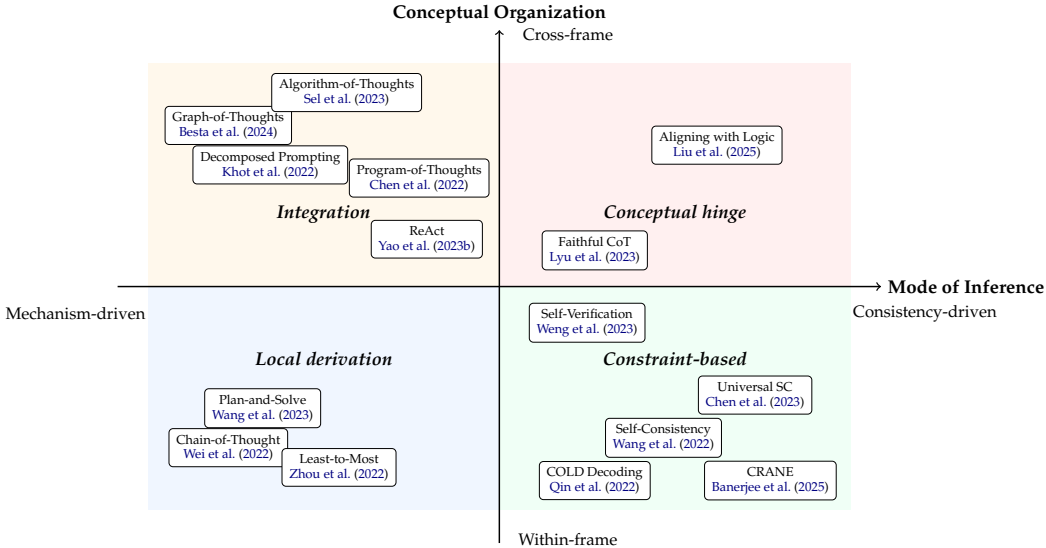
\begin{figure}[ht!]
\centering
\resizebox{\linewidth}{!}{%
\begin{tikzpicture}[x=1.45cm,y=1.25cm]

\definecolor{mechsingle}{RGB}{220,235,255}
\definecolor{mechmulti}{RGB}{255,240,210}
\definecolor{consingle}{RGB}{220,255,235}
\definecolor{conmulti}{RGB}{255,220,220}

\fill[mechsingle, opacity=0.42] (-4.6,-3.4) rectangle (0,0);
\fill[mechmulti,  opacity=0.42] (-4.6,0) rectangle (0,3.4);
\fill[consingle,  opacity=0.42] (0,-3.4) rectangle (4.6,0);
\fill[conmulti,   opacity=0.42] (0,0) rectangle (4.6,3.4);

\draw[->, thick] (-5,0) -- (5,0) node[right] {\bfseries Mode of Inference};
\draw[->, thick] (0,-3.9) -- (0,3.9) node[above] {\bfseries Conceptual Organization};

\node[anchor=east] at (-4.55,-0.40) {\small Mechanism-driven};
\node[anchor=west] at (4.55,-0.40) {\small Consistency-driven};

\node[anchor=north] at (0.9,-3.65) {\small Within-frame};
\node[anchor=south] at (0.9,3.65) {\small Cross-frame};

\node[font=\bfseries\itshape] at (-2.3,-1.10) {Local derivation};
\node[font=\bfseries\itshape] at ( 2.3,-1.10) {Constraint-based};
\node[font=\bfseries\itshape] at (-2.3, 1.10) {Integration};
\node[font=\bfseries\itshape] at ( 2.3, 1.10) {Conceptual hinge};

\tikzset{
  method/.style={
    draw,
    rounded corners=2.2pt,
    fill=white,
    align=center,
    font=\scriptsize,
    inner xsep=4pt,
    inner ysep=3pt
  }
}

\node[method] at (-3.55,-2.45) {Chain-of-Thought\\\cite{wei2022chain-of-thought}};
\node[method] at (-2.10,-2.75) {Least-to-Most\\\cite{zhou2022least-to-most}};
\node[method] at (-3.10,-1.85) {Plan-and-Solve\\\cite{wang2023plan-and-solve}};

\node[method] at (-3.55, 2.45) {Graph-of-Thoughts\\\cite{besta2024graph-of-thoughts}};
\node[method] at (-2.00, 2.95) {Algorithm-of-Thoughts\\\cite{sel2023algorithm-of-thoughts}};
\node[method] at (-3.00, 1.85) {Decomposed Prompting\\\cite{khot2022decomposed-prompting}};
\node[method] at (-1.05, 1.65) {Program-of-Thoughts\\\cite{chen2022program-of-thoughts}};
\node[method] at (-0.95, 0.72) {ReAct\\\cite{yao2023react}};

\node[method] at ( 2.15,-2.3) {Self-Consistency\\\cite{wang2022self-consistency}};
\node[method] at ( 3.35,-1.65) {Universal SC\\\cite{chen2023universal}};
\node[method] at ( 1.15,-0.55) {Self-Verification\\\cite{weng2023self-verification}};
\node[method] at ( 1.25,-2.95) {COLD Decoding\\\cite{qin2022cold-decoding}};
\node[method] at ( 3.55,-2.95) {CRANE\\\cite{banerjee2025crane}};

\node[method] at (1.25,0.55) {Faithful CoT\\\cite{lyu2023faithful}};
\node[method] at (2.85,2.15) {Aligning with Logic\\\cite{liu2025aligning-with-logic}};

\end{tikzpicture}%
}
\caption{
Prior work positioned in the reasoning-regime phase diagram.
}
\label{fig:related-work-phase-diagram}
\end{figure}

Prior work on LLM reasoning naturally clusters along two dimensions, 
which motivates the axes of our reasoning geometry. The first is 
whether a method strengthens reasoning by making intermediate mechanisms 
explicit or by enforcing global consistency. The second is whether it 
operates within a single representational structure or coordinates 
multiple structures. Figure~\ref{fig:related-work-phase-diagram} shows 
the placement of representative methods; detailed discussion is in 
Appendix~\ref{app:prior work reasoning regimes}. This clustering motivates four reasoning regimes that we use to 
categorize both prior methods and our twelve problems:
\begin{itemize}
    \item \textbf{Local derivation} (Q1, Q2, Q5, Q7): mechanism-driven, within-frame.
    \item \textbf{Integration} (Q4, Q9, Q12): mechanism-driven, cross-frame.
    \item \textbf{Constraint-based} (Q3, Q8, Q10): consistency-driven, within-frame.
    \item \textbf{Conceptual hinge} (Q6, Q11): consistency-driven, cross-frame.
\end{itemize}

\section{Scoring Criteria}
\label{sec:scoring criteria}
Evaluating reasoning in highly abstract theoretical domains is inherently nontrivial.
In string theory and quantum field theory, correctness is not a binary notion:
a response may state the correct conclusion while omitting essential intermediate structure,
or reproduce formal manipulations without respecting conceptual constraints. To resolve this ambiguity and capture distinct failure modes of reasoning,
we introduce a five-level grading scheme that decomposes correctness
into progressively deeper dimensions of reasoning reconstruction.
For each level, we provide a representative example together with
a brief explanation of the corresponding scoring decision. 


\textbf{Level 0: Statement Correctness.} The final claim is factually correct and free of fatal conceptual errors,
even if the reasoning leading to it is absent or incorrect. Almost all evaluated
models meet this level (see Section~\ref{subsec:overal performance}),
reflecting that Level~0 probes recall rather than reasoning reconstruction.
An example is provided in Appendix~\ref{app:additionalexample}.

\textbf{Level 1: Key Concept Awareness.} 
    The response identifies the relevant concept, theorem, mechanism, or structural principle underlying the statement, even if the explanation remains superficial and non-reasoning.

\noindent\textit{Success Example (Q6, Gemini-3.1-Pro-Preview)}:

\begin{tcolorbox}[modelout]``To understand why $(5+1)$-dimensional superconformal field theories (SCFTs) are classified as intrinsically relative theories ... concept of a relative quantum field theory --- formalized by Freed, Moore, Segal, Witten, and Tachikawa --- describes a theory that does not possess a canonical, single-valued partition function. Instead, its partition function is a vector residing in a Hilbert space defined by a higher-dimensional topological quantum field theory (TQFT) ...''
\end{tcolorbox}

\noindent\textit{Explanation:} The response explicitly identifies the structural notion of a
\emph{relative quantum field theory}, its \emph{partition function} and related \emph{topological quantum field theory}, which are key concepts
underlying the statement being tested. Therefore it satisfies Level~1.

\textbf{Level 2: Reasoning Chain Presence.} 
Level~2 distinguishes the presence of relevant ingredients from a
causal explanatory structure linking them. A success example is provided in Appendix~\ref{app:additionalexample}.

\noindent\textit{Failure example (Q10, Gemini-2.5-Flash)}:

\begin{tcolorbox}[modelout]
``For a single D-brane ($p$-brane), open strings ending on it give rise to a $U(1)$ gauge theory on its $(p+1)$-dimensional worldvolume. Naively, this $U(1)$ field might be thought of as conventional electromagnetism ... requiring it to provide a $\mathrm{Spin}^c$ structure on the D-brane worldvolume ... 1. D-brane Charges and K-theory: ... 2. The Role of the NS-NS $B$-field: ... 3. Challenge to a Conventional $U(1)$: ... 4. The Spin$^c$ Structure Resolution: ... first Chern class $c_1(L)$ of this line bundle $L$ is given by $c_1(L) = \frac{1}{2\pi} F$, where $F$ is the curvature of the connection $A$ ... For these fermions to be consistently defined globally on the D-brane worldvolume, the worldvolume must admit a $\mathrm{Spin}^c$ structure..."
\end{tcolorbox}

\textit{Explanation:} The response mentions several concepts, such as K-theory, globally defined fermions 
and the $U(1)$ gauge field, which are relevant ingredients to reconstruct the statement in the question, thus satisfies the Level 1. However, it does not provide a 
reasoning chain explaining why the interplay between these concepts implies the existence of a 
Spin$^c$ structure. The conclusion therefore appears as an unsupported statement 
rather than the result of a coherent derivation, and thus fails Level~2.

\textbf{Level 3: Tacit Step Reconstruction.} 
   Intermediate reasoning steps that are typically omitted in textbook or research literature
are explicitly reconstructed and integrated into the argument. A success example is provided in Appendix~\ref{app:additionalexample}.


\noindent\textit{Failure example (Q3, GPT-4.1)}:

\begin{tcolorbox}[modelout]``Let's analyze why, in $(2+1)$-dimensional $U(1)$ Chern-Simons theory at level $k$, only when $k$ is even can the theory be bosonic ... bosonic theory is one that can be consistently defined on any oriented 3-manifold, without requiring a spin structure ... the theory is bosonic if all local excitations have integer spin ... For the minimal charge $q=1$, the spin is $h_1 = \frac{1}{2k}$ ... The Wilson line of charge $k$ (i.e., $q=k$) has spin: ... For this to be an integer (i.e., bosonic), $k$ must be even ... If $k$ is odd, $h_k$ is half-integer, so the Wilson line of charge $k$ is a fermion. This means the theory requires a spin structure and is thus fermionic ..."\end{tcolorbox}

\noindent\textit{Explanation:} Although the response correctly states that if a theory is bosonic, all of its local excitations have integer spin, and show when $k$ is odd, there are half-integer, i.e., fermionic anyons.
However, it misses the key tacit step that only excitations with trivial mutual braiding
with all other anyons can be regarded as genuinely local/transparent excitations.
The answer therefore contains the relevant ingredients and a partial reasoning chain, satisfying Level 2, 
but misses the tacit step required for Level~3.

\textbf{Level 4: Enrichment.} 
Level~4 is intended as a bonus level. Beyond reconstructing the tacit reasoning, the response demonstrates a broader conceptual awareness of the structure underlying the statement. This may include clarifying the regime of validity, discussing conceptual
boundaries, presenting alternative perspectives, or illustrating the argument through explicit examples. 

\noindent\textit{Success example (Q4, Gemini-3.1-Pro-Preview)}:
\begin{tcolorbox}[modelout]``Physical Evidence: This theoretical framework is heavily supported by Lattice QCD simulations ... At low temperatures, the system is in an unbroken phase, and Wilson loops obey the area law. At a critical high temperature $T_c$, the system undergoes a phase transition where the $\mathbb{Z}_N$ symmetry breaks spontaneously, Wilson loops transition to a perimeter law, and the theory enters a deconfined phase (analogous to the Quark-Gluon Plasma experimentally observed in heavy-ion collisions at the LHC and RHIC ..."\end{tcolorbox}

\noindent\textit{Explanation:}
This response goes beyond reconstructing the core reasoning by providing more explicit physical pheonomenon ,i.e. phase transition of lattice gauge theories governd by the main mechanism reconstructed in the question. Such an illustrative viewpoint provides a better understanding of this tacit knowledge, and qualifies as a Level~4 bonus insight.

Detailed evaluation criteria for each question are provided in 
Appendix~\ref{app:eval_criteria}.



\section{Evaluation}


\subsection{Overall Performance}\label{subsec:overal performance}
Table~\ref{tab:overall} summarizes model performance across the five evaluation
levels, together with the total score across all levels.

\begin{table}[ht!]
\centering
\small
\setlength{\tabcolsep}{4pt}
\renewcommand{\arraystretch}{1.1}
\resizebox{\textwidth}{!}{%
\begin{tabular}{lccccc|c}
\toprule
 & L0 (Statement) & L1 (Result) & L2 (Reasoning) & L3 (Tacit) & L4 (Enrichment) & Overall \\
\midrule

Gemini-2.5-flash
& \cellcolor{green!35}1.000
& \cellcolor{green!15}0.833
& \cellcolor{green!15}0.750
& \cellcolor{red!30}0.083
& \cellcolor{red!40}0.000
& \cellcolor{yellow!25}2.666 \\

Gemini-3.1-pro-preview
& \cellcolor{green!35}1.000
& \cellcolor{green!35}1.000
& \cellcolor{green!35}1.000
& \cellcolor{green!25}0.917
& \cellcolor{yellow!25}0.500
& \cellcolor{green!25}4.417 \\

GPT-5.2
& \cellcolor{green!35}1.000
& \cellcolor{green!35}1.000
& \cellcolor{green!25}0.917
& \cellcolor{yellow!35}0.583
& \cellcolor{red!30}0.083
& \cellcolor{green!15}3.583 \\

GPT-4.1
& \cellcolor{green!35}1.000
& \cellcolor{green!35}1.000
& \cellcolor{green!25}0.917
& \cellcolor{red!20}0.250
& \cellcolor{red!40}0.000
& \cellcolor{yellow!35}3.167 \\

Deepseek-V3.2 (non-thinking)
& \cellcolor{green!35}1.000
& \cellcolor{green!35}1.000
& \cellcolor{yellow!35}0.667
& \cellcolor{red!20}0.250
& \cellcolor{red!40}0.000
& \cellcolor{yellow!35}2.917 \\

Deepseek-V3.2 (thinking)
& \cellcolor{green!35}1.000
& \cellcolor{green!25}0.917
& \cellcolor{green!15}0.750
& \cellcolor{red!20}0.250
& \cellcolor{red!30}0.083
& \cellcolor{yellow!35}3.000 \\

Kimi-K2-thinking
& \cellcolor{green!35}1.000
& \cellcolor{green!15}0.833
& \cellcolor{green!15}0.750
& \cellcolor{red!20}0.250
& \cellcolor{red!40}0.000
& \cellcolor{yellow!35}2.833 \\

Qwen3.5-397b
& \cellcolor{green!25}0.917
& \cellcolor{green!15}0.833
& \cellcolor{green!15}0.833
& \cellcolor{yellow!25}0.500
& \cellcolor{red!20}0.250
& \cellcolor{yellow!35}3.333 \\

Minimax-m2.7
& \cellcolor{green!25}0.917
& \cellcolor{green!15}0.750
& \cellcolor{yellow!35}0.583
& \cellcolor{red!20}0.167
& \cellcolor{red!40}0.000
& \cellcolor{yellow!25}2.417 \\

Nemotron-3-super
& \cellcolor{green!25}0.917
& \cellcolor{yellow!25}0.500
& \cellcolor{yellow!25}0.417
& \cellcolor{red!20}0.250
& \cellcolor{red!40}0.000
& \cellcolor{yellow!25}2.084 \\

\bottomrule
\end{tabular}}
\caption{Average success rate across 12 problems at each evaluation level. Colors indicate performance from low (red) to high (green) using a uniform scale, and the Overall column is colored on the same scale normalized by the maximum possible score of~5 (i.e., Overall\,/\,5).}
\label{tab:overall}
\end{table}

First, we observe a clear stratification across evaluation levels.
All models achieve near-saturated performance at Levels~0--2,
corresponding to statement correctness, identification of relevant concepts,
and the presence of a basic reasoning chain. In contrast, performance exhibits a sharp qualitative change at deeper levels.
Level~3 introduces a strong separation between models.
While leading systems such as Gemini-3.1-pro-preview maintain high performance
($\sim 0.92$), most other models show substantial degradation,
typically dropping to the range $0.17$--$0.50$.
This indicates that reconstructing omitted intermediate reasoning steps
is a significant challenge even when the final result is correctly identified.

Level~4 (Enrichment) should be interpreted differently.
This level is designed as a bonus level that probes broader conceptual awareness,
such as identifying regimes of validity or presenting alternative perspectives.
As expected, most models achieve low scores at this level.
We therefore do not treat Level~4 as a primary indicator of failure,
but rather as an additional signal of advanced reasoning behavior.

Performance varies moderately across research areas (see 
Appendix~\ref{app:domain}); reasoning type is the primary 
source of difficulty, as we show next.

\subsection{Reasoning Geometry}

We now analyze model performance across the four reasoning regimes
introduced in Section~2.
Table~\ref{tab:task_summary} reports the average score obtained by each model
within each regime after aggregating across evaluation levels. This view provides a coarse comparison of task difficulty and model capability without exposing the internal structure of the grading scheme.


\begin{table}[ht!]
\centering
\small
\setlength{\tabcolsep}{4pt}
\renewcommand{\arraystretch}{1.1}
\begin{tabular}{lcccc}
\toprule
Model & Local derivation & Integration & Constraint-based & Conceptual hinge\\
\midrule

Gemini-2.5-flash
& \cellcolor{yellow!35}3.250
& \cellcolor{yellow!25}2.334
& \cellcolor{yellow!25}2.334
& \cellcolor{yellow!25}2.500 \\

Gemini-3.1-pro-preview
& \cellcolor{green!25}4.500
& \cellcolor{green!25}4.667
& \cellcolor{green!15}4.000
& \cellcolor{green!25}4.500 \\

GPT-5.2
& \cellcolor{green!15}3.750
& \cellcolor{green!15}3.667
& \cellcolor{yellow!35}3.334
& \cellcolor{green!15}3.500 \\

GPT-4.1
& \cellcolor{green!15}3.500
& \cellcolor{yellow!35}3.333
& \cellcolor{yellow!35}3.000
& \cellcolor{yellow!35}3.000 \\

Deepseek-V3.2 (non-thinking)
& \cellcolor{yellow!35}3.000
& \cellcolor{yellow!35}3.000
& \cellcolor{yellow!25}2.666
& \cellcolor{yellow!25}2.500 \\

Deepseek-V3.2 (thinking)
& \cellcolor{green!15}3.500
& \cellcolor{yellow!25}2.667
& \cellcolor{yellow!35}3.000
& \cellcolor{yellow!25}2.500 \\

Kimi-K2-thinking
& \cellcolor{green!15}3.750
& \cellcolor{yellow!35}3.000
& \cellcolor{yellow!35}3.000
& \cellcolor{red!20}1.000 \\

Qwen3.5-397b
& \cellcolor{green!15}3.750
& \cellcolor{green!15}3.667
& \cellcolor{green!15}3.667
& \cellcolor{red!30}0.500 \\

Minimax-m2.7
& \cellcolor{yellow!35}2.750
& \cellcolor{yellow!25}2.000
& \cellcolor{yellow!25}2.667
& \cellcolor{red!30}0.500 \\

Nemotron-3-super
& \cellcolor{yellow!25}2.500
& \cellcolor{yellow!25}2.000
& \cellcolor{yellow!25}2.667
& \cellcolor{red!20}1.000 \\

\bottomrule
\end{tabular}
\caption{
Average model performance by reasoning regime. For each task category, scores are first summed across the five evaluation levels (L0--L4) for each question, and then averaged across all questions belonging to that reasoning regime. Colors indicate performance from low (red) to high (green) on a uniform scale (score\,/\,5).
}
\label{tab:task_summary}
\end{table}

Two qualitative observations emerge immediately.
First, mechanism-driven regimes (Local Derivation and Integration)
consistently exhibit higher performance across models.
Second, performance degrades systematically as we move toward
consistency-driven regimes, with the most severe drop occurring
in Conceptual Hinge tasks.

\begin{figure*}[ht!]
\centering
\includegraphics[width=\textwidth]{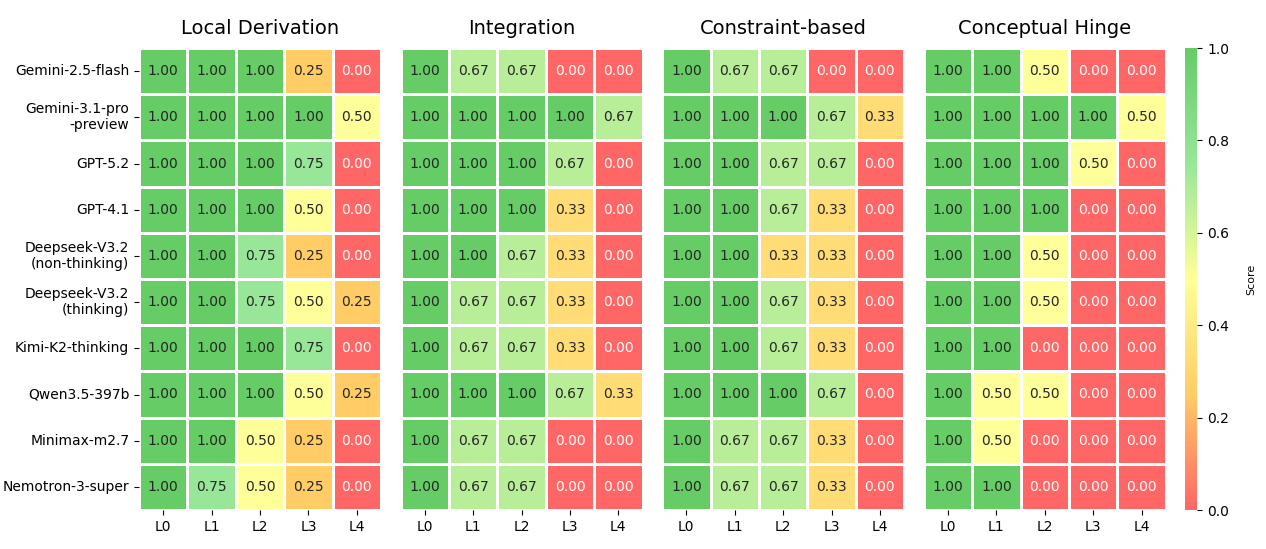}
\caption{
Model performance across four reasoning regimes and five cumulative evaluation levels.
Each panel corresponds to one reasoning regime.
Columns represent evaluation Levels~0--4.
Rows correspond to evaluated language models.
Cell color indicates the average score within each regime-level combination. 
}
\label{fig:reasoning_heatmap}
\end{figure*}

This pattern indicates that the primary source of difficulty
is not domain knowledge, but the nature of the reasoning process itself. Figure~\ref{fig:reasoning_heatmap} summarizes model performance across the four reasoning regimes and five evaluation levels. Detailed tables for each reasoning tasks are included in Appendix~\ref{app:reasoning geometry}.

\noindent\textbf{Local derivation tasks.}
In the mechanism-driven and single-structure quadrant,
performance is near-saturated at Levels~0--2 across all models,
indicating that explicit derivations within a stable conceptual frame
are reliably handled.
Separation begins at Level~3, where models must reconstruct tacit
intermediate steps, but degradation remains moderate (Table~\ref{tab:local_derivation}).
This suggests that models are effective at sequentially expanding
well-structured reasoning chains when the underlying representation
is fixed.

\noindent\textbf{Integration tasks.} 
When remaining mechanism-driven but requiring cross-domain synthesis, performance begins to differentiate.
While Levels~0--2 remain relatively strong,
substantial variation appears at Level~3 (Table~\ref{tab:integration}).
This indicates that the primary difficulty lies not in executing
individual derivations, but in combining distinct reasoning threads
into a coherent argument.
Nevertheless, performance remains significantly higher than in
consistency-driven regimes, suggesting that the conceptual frame
remains largely stable .

\noindent\textbf{Constraint-based reasoning tasks.}
In the consistency-driven but single-structure quadrant,
performance degradation appears earlier.
Even at Level~2 (Table~\ref{tab:constraint}), several models show noticeable decline,
indicating that enforcing global consistency conditions introduces
additional reasoning complexity.
At Level~3, models diverge sharply:
while stronger systems can propagate constraints through the argument,
weaker models often fail to complete the tacit reconstruction.
This suggests that recognizing a constraint is insufficient;
the difficulty lies in integrating it consistently into the reasoning process .

\noindent\textbf{Conceptual hinge tasks.}
The most substantial degradation occurs in the quadrant combining
consistency-driven inference with multi-structure organization (Table~\ref{tab:hinge}).
Performance drops already at Level~2 for many models,
and collapses almost entirely at Level~3.
These tasks require identifying a latent structural distinction
that resolves an apparent conceptual tension before any derivation proceeds.
Unlike other regimes, failure occurs not in executing reasoning steps,
but in initiating the correct representational frame.

\subsection{Failure Analysis}
\noindent\textbf{Hypothesized failure mechanism.}
The observed pattern suggests that current LLMs predominantly operate
in a forward-expansion regime, in which locally coherent reasoning
chains are extended within a fixed representational frame.

In contrast, tasks in the upper-right quadrant require an initial
representational shift.
Before any derivation proceeds, the model must identify a latent
structural distinction that reorganizes the interpretation of the problem
(e.g., distinguishing different anomaly types in Q11 or recognizing
relative vs. absolute formulations in Q6).

Our results indicate that this restructuring step is not reliably triggered.
Failures therefore arise not from lack of technical knowledge,
but from difficulty in autonomously reorganizing the conceptual space
under global consistency constraints.

\noindent\textbf{Prompt sensitivity of conceptual hinge: a case study of Q11.}
To further probe the nature of failures in conceptual hinge tasks,
we conduct a perturbation experiment on a representative problem (Q11).
We compare the original formulation (Version~A) with modified prompts
(Versions~B and~C).
The modified prompts retain the same core question as Version~A,
but include additional hinting sentences at the end:

\begin{tcolorbox}[qcard]
\textbf{Q11B.} ...Resolve this contradiction. In answering, pay particular attention to whether the word ``anomaly'' is being used in exactly the same sense in the two statements.
\end{tcolorbox}

\begin{tcolorbox}[qcard]
\textbf{Q11C.} ...Resolve this contradiction by carefully distinguishing different notions of anomaly.
\end{tcolorbox}

Importantly, these modifications do not introduce new technical content.
Instead, they make explicit a structural distinction that is implicit
in the original formulation.

The behavior of models across the three versions is summarized in
Table~\ref{tab:q11_total}, where scores are reported as the total number
of evaluation levels satisfied (i.e., the sum over L0--L4).
\begin{table}[ht!]
\centering
\begin{tabular}{lccc}
\toprule
Model & Q11 & Q11B & Q11C \\
\midrule
Gemini2.5-flash 
& \cellcolor{red!40}1 & \cellcolor{red!20}2 & \cellcolor{red!20}2 \\
Gemini-3.1-pro-preview 
& \cellcolor{green!25}4 & \cellcolor{green!25}4 & \cellcolor{green!25}4 \\
GPT-5.2 
& \cellcolor{green!25}4 & \cellcolor{green!25}4 & \cellcolor{green!35}5 \\
GPT-4.1 
& \cellcolor{yellow!35}3 & \cellcolor{red!40}1 & \cellcolor{red!20}2 \\
Deepseek-V3.2 (non-thinking) 
& \cellcolor{red!20}2 & \cellcolor{yellow!35}3 & \cellcolor{yellow!35}3 \\
Deepseek-V3.2 (thinking) 
& \cellcolor{red!40}1 & \cellcolor{red!20}2 & \cellcolor{red!20}2 \\
Kimi-K2-thinking 
& \cellcolor{red!40}1 & \cellcolor{red!20}2 & \cellcolor{red!20}2 \\
Qwen3.5-397b 
& \cellcolor{red!60}0 & \cellcolor{green!25}4 & \cellcolor{red!20}2 \\
Minimax-m2.7 
& \cellcolor{red!40}1 & \cellcolor{red!20}2 & \cellcolor{red!40}1 \\
Nemotron-3-super 
& \cellcolor{red!40}1 & \cellcolor{red!20}2 & \cellcolor{red!20}2 \\
\hline
\end{tabular}
\caption{Total scores (L0--L4 summed) for Q11 under different prompt variants.}
\label{tab:q11_total}
\end{table}
In the original formulation (Version~A), most models fail to reach
Level~3, indicating an inability to reconstruct the tacit reasoning
required to resolve the conceptual tension.
Under the modified prompts, however, several models exhibit
substantial improvement.
For example, Qwen3.5-397b improves from a score of $0$ to $4$
under Version~B, representing a transition from complete failure
to near-complete reasoning reconstruction.
Similarly, Deepseek-V3.2 (non-thinking) reaches Level~3 under
modified prompts, indicating successful reconstruction of tacit
intermediate steps.
GPT-5.2 further improves from Level~3 to Level~4 under Version~C,
demonstrating that enriched reasoning can be triggered under
appropriate formulations.

Notably, for all evaluated models except for GPT-4.1, performance under the modified
prompts is never worse than the original formulation, and is often
substantially higher.
This monotonic behavior suggests that the additional hinting sentences
act as a trigger for latent reasoning capabilities, rather than
introducing new information.

At the same time, performance across Versions~B and~C is not stable.
The same model may achieve different reasoning levels under
alternative formulations, indicating sensitivity to prompt phrasing.
This suggests that while conceptual hinge reasoning can be elicited,
it is not robustly activated.




\section{Related Work}

\textbf{LLM Reasoning benchmarks.}
LLM reasoning evaluation has made steady progress by raising the
difficulty ceiling, from broad multitask
knowledge~\citep{hendrycks2021mmlu} to research-frontier problems designed
to resist all current
models~\citep{glazer2024frontiermath,pano2025hle}.
The problems range from grade-school
arithmetic~\citep{cobbe2021gsm8k} and competition-level
mathematics~\citep{hendrycks2021math,gao2024omnimath} to graduate-level
expert questions~\citep{rein2024gpqa}.
However, current benchmarks primarily evaluate correctness as
final-answer matching or forced-choice selection, a design that cannot
distinguish valid inferential paths from pattern-matched or
step-skipping ones.
Gains on answer-matching tasks do not reliably transfer to tasks
requiring structured intermediate
reasoning~\citep{srivastava2022bigbench,plaat2025multi}.
This limitation is particularly acute in research-level quantum field theory and string
theory: experts routinely suppress entire inferential moves by
convention, multiple non-equivalent derivation paths coexist, and the
evaluatively interesting question is whether the model has reconstructed
the reasoning behind its answer.

\textbf{Reasoning evaluation in steps.}
Recognizing that outcome accuracy alone is inadequate, recent work has
also explored evaluating intermediate reasoning steps.
One direction assigns correctness and quality labels to individual
reasoning steps~\citep{lightman2023process,xia2024reasoneval}.
Another investigates when LLM judges can find agreement on
well-defined step-level reasoning
tasks~\citep{sawada2023arb,zheng2023llmjudge}.
A third engages domain experts in live dialogue with models, where
targeted follow-up questions reveal reasoning failures that static
evaluation cannot
detect~\citep{frieder2024mathematical,collins2024interactive}.
All of these efforts primarily presuppose an explicit chain and ask
whether each link is valid.
They cannot assess whether a link is missing.
In expert theoretical physics, missing links are the norm: practitioners
routinely leave reasoning unstated because the community treats it as
obvious, and this tacit knowledge is precisely what separates competent
derivation from surface-level answer production.
Our five-level rubric targets this distinction by evaluating the
reconstruction of suppressed steps and implicit constraints through
expert annotation.

\textbf{Physics and high-energy-theory reasoning.}
Physics reasoning requires grounding in physical law and multi-step
symbolic manipulation under domain-specific constraints; at the research
level, it further demands navigation of deeply abstract formalisms where
the reasoning structure itself is convention-dependent.
Recent physics
benchmarks~\citep{wang2024scibench,xu2025ugphysics,qiu2025phybench,zhang2025physreason,zheng2025physics,zhang2025abench}
have introduced partial-credit scoring and solution-step decomposition,
but all still require a canonical correct answer, a property that
research-level quantum field theory and string theory questions do not possess.
Work on LLMs in research-level theoretical physics has shown that models
can follow supplied derivational templates~\citep{pan2025manybody} but
struggle with open-ended research problems, where symbolic and logical
errors dominate~\cite{chung2025tpbench}.
Domain adaptation can capture terminological
conventions~\citep{richmond2025feyntune}, and targeted fine-tuning can
install narrow calculational skills~\citep{cai2025lac}.
Different from these efforts, our rubric and expert-curated dataset
directly evaluate whether models can reconstruct the implicit reasoning
of expert practitioners: the suppressed steps, unstated assumptions, and
representation choices that define tacit inferential competence in quantum field theory
and string theory.

\section{Conclusion}
We constructed a small expert-curated dataset of twelve questions in 
quantum field theory and string theory, and introduced a five-level 
rubric to evaluate tacit reasoning reconstruction. Models perform 
near-ceiling on explicit derivations but fail systematically at 
Level~3, with failures concentrated in conceptual hinge tasks. The 
prompt perturbation experiment shows the bottleneck is representation 
selection rather than missing knowledge. We argue that highly 
abstract theoretical physics provides a uniquely sensitive testbed for 
exposing the epistemic limits of current evaluation paradigms.


\bibliography{colm2026_conference}
\bibliographystyle{colm2026_conference}

\appendix
\section{Dataset for tacit knowledge}
\label{app:dataset}
We construct a small expert-curated set of twelve questions drawn from quantum field theory and string theory. 
\begin{tcolorbox}[qcard]
\textbf{Q1.} In (1+1)-dimensional quantum field theories, why cannot continuous global symmetries be spontaneously broken?
\end{tcolorbox}

\begin{tcolorbox}[qcard]
\textbf{Q2.} In (1+1)-dimensional conformal field theories, why 2-point functions for primary operators must be power functions?
\end{tcolorbox}

\begin{tcolorbox}[qcard]
\textbf{Q3.} For the (2+1)-dimensional $U(1)$ Chern-Simons theory at level $k$, why only when $k$ is even can the theory be bosonic?
\end{tcolorbox}

\begin{tcolorbox}[qcard]
\textbf{Q4.} It is often stated that the following three concepts are closely related to each other: 1. confined/deconfined phase of a gauge theory 2. surface/perimeter law of the correlation function of Wilson lines 3. preserving/breaking of the 1-form symmetry. Provide the reasoning for why they are connected.
\end{tcolorbox}

\begin{tcolorbox}[qcard]
\textbf{Q5.} In (3+1)-dimensional $\mathcal{N}=1$ supersymmetric field theory, why supersymmetric vacua are in one-to-one correspondence
with the zero’s of the scalar potential?
\end{tcolorbox}

\begin{tcolorbox}[qcard]
\textbf{Q6.} It is often stated that (5+1)-dimensional superconformal field theories are ``intrinsically" relative theories. Explain why they are relatvie and why their relativeness is intrinsic.
\end{tcolorbox}

\begin{tcolorbox}[qcard]
\textbf{Q7.} It is stated that T-duality in string theory exchanges Kaluza–Klein momentum modes with string winding modes, without supplying the underlying worldsheet derivation. Reconstruct the worldsheet reasoning that explains why T-duality exchanges momentum and winding modes.
\end{tcolorbox}

\begin{tcolorbox}[qcard]
\textbf{Q8.} Superstring theories require GSO projections to be consistent, which is often stated from a conformal field theory perspective as the requirement of modular invariance. Explain the reasoning why modular-invariance is required for string theory and how is this related to GSO projection.
\end{tcolorbox}

\begin{tcolorbox}[qcard]
\textbf{Q9.} The gauge algebras leaving on the worldvolume of D-branes are usually argued to be realized by Chan-Paton factors on the open string boundary. Explain how these gauge algebras, including u(n), so(n) and sp(n) can be read explicitly from physical degrees of freedom as boundary Majorana fermions on the open string worldsheet. 
\end{tcolorbox}

\begin{tcolorbox}[qcard]
\textbf{Q10.} For a single D-brane in type II string theory, naively it is endowed with a $U(1)$ gauge field. However, it is usually claimed that the precise characterization of this $U(1)$ field is not a conventional one, but providing a Spin$^c$ structure on the brane worldvolume. Explain why this is the case.
\end{tcolorbox}

\begin{tcolorbox}[qcard]
\textbf{Q11.} One common statement is that if a global symmetry suffers from an anomaly, then it is preserved at the classical level but not at the quantum level. Another common statement is that one can use the global symmetry and its anomaly to match the ultraviolet and the infrared physics along the renormalization group flow. Given the renormalization group flow is in general quantum mechanical, these two statements seem to be a contradiction. Resolve this contradiction. 
\end{tcolorbox}

\begin{tcolorbox}[qcard]
\textbf{Q12.} It is often stated that the symmetry breaking can only happen in infinite-volume space. Explain the reasoning behind this general yoga.
\end{tcolorbox}

\section{Positioning prior work in the reasoning-regime phase diagram}
\label{app:prior work reasoning regimes}

\paragraph{Mechanism-driven + single-structure.}
A first line of work improves reasoning by making intermediate derivations more explicit while keeping the task within a shared conceptual frame. Chain-of-Thought prompting elicits step-by-step reasoning traces for multi-step inference \citep{wei2022chain-of-thought}. Least-to-Most prompting further decomposes a problem into an ordered sequence of simpler subproblems \citep{zhou2022least-to-most}, while Plan-and-Solve separates planning from execution to reduce missing-step errors \citep{wang2023plan-and-solve}. Tree-of-Thoughts extends this family by allowing branching exploration and backtracking over intermediate thoughts \citep{yao2023tree-of-thoughts}. Although these methods differ in search breadth and control flow, they remain primarily mechanism-driven and largely preserve a single task representation.

\paragraph{Mechanism-driven + multi-structure.}
Shifting along the structural axis, another line of work still centers on explicit mechanisms but distributes reasoning across multiple interacting representations, modules, or action states. Graph-of-Thoughts generalizes chain- and tree-based reasoning to graph-structured dependencies among reasoning units \citep{besta2024graph-of-thoughts}, and Algorithm-of-Thoughts introduces algorithmically organized exploration over intermediate states \citep{sel2023algorithm-of-thoughts}. Program-of-Thoughts separates natural-language reasoning from executable computation \citep{chen2022program-of-thoughts}, while Decomposed Prompting organizes complex tasks into modular subproblems handled by specialized prompts or components \citep{khot2022decomposed-prompting}. ReAct further couples reasoning with external actions and observations, so interaction with an environment becomes part of the evolving inference process \citep{yao2023react}. Relative to the previous regime, the main shift here is not simply longer derivation, but coordination across heterogeneous intermediate forms.

\paragraph{Consistency-driven + single-structure.}
A complementary literature studies reasoning through global agreement and admissibility rather than the quality of any single derivation. Self-Consistency aggregates multiple sampled reasoning paths to select the most consistent answer \citep{wang2022self-consistency}, Universal Self-Consistency extends this idea beyond settings with easily extractable final answers \citep{chen2023universal}, and Self-Verification checks candidate conclusions by reasoning backward from them to the original problem \citep{weng2023self-verification}. In parallel, constrained-generation methods such as COLD Decoding~\citep{qin2022cold-decoding} and CRANE~\citep{banerjee2025crane} impose restrictions on the generation space itself. While these approaches differ in where consistency is enforced---across sampled outputs, through backward checking, or during decoding---they largely evaluate candidate solutions within a single representational frame.

\paragraph{Consistency-driven + multi-structure.}
The final regime emphasizes consistency across transformed or formally distinct views of the same problem. Faithful Chain-of-Thought~\citep{lyu2023faithful} maps natural-language problems into symbolic reasoning chains that are executed by a deterministic solver, making correctness depend on agreement between linguistic and formal representations. Aligning with Logic~\citep{liu2025aligning-with-logic} studies whether model judgments remain stable under explicit logical transformations, such as transitivity, commutativity, and negation invariance. Compared with the previous regime, these methods move beyond consistency over candidate answers alone and instead ask whether conclusions remain coherent across multiple representations or transformed views of the same underlying reasoning problem.

\section{Evaluation Criteria}

\subsection{Additional examples}
\label{app:additionalexample}
\paragraph{Level 0: Statement Correctness.}
\noindent\textit{Example (Q8, GPT-5.2)}:
\begin{tcolorbox}[modelout]
``In perturbative string theory the basic object is the sum over inequivalent worldsheet geometries ... the requirement that we are summing over *inequivalent* tori forces **modular invariance**. The **GSO projection** is then the mechanism (in superstrings) that removes states so that the torus partition function becomes modular invariant ...''
\end{tcolorbox}
\noindent\textit{Explanation:} The response correctly state the modular invariance as a requirement of consistent string theory, and point out GSO projection leads to the modular invariance. Therefore it satisfies level 0. 

\paragraph{Level 2: Reasoning Chain Presence.}
\noindent\textit{Success example (Q10, Deepseek-V3.2 thinking)}.

\begin{tcolorbox}[modelout]``In type II string theory, a single D$p$-brane naturally carries a $U(1)$ gauge field $A$ on its $(p+1)$-dimensional worldvolume $M$. However, this $U(1)$ field is not merely a conventional gauge connection; it is intrinsically tied to the global topological structure required for consistency of the worldvolume fermions ... On an oriented Riemannian manifold $M$, a spin structure exists if and only if the second Stiefel-Whitney class $w_2(TM) \in H^2(M, \mathbb{Z}_2)$ vanishes. However, many manifolds of interest in string theory (e.g., complex submanifolds of Calabi-Yau manifolds) may not satisfy $w_2(TM) = 0$ ... In such cases, a weaker condition is that $M$ admits a Spin$^c$ structure, which is a lift of the structure group ... Spin$^c$ structure exists if and only if $w_2(TM)$ is the mod 2 reduction of an integral cohomology class $c \in H^2(M, \mathbb{Z})$. This condition is equivalent to the existence of a complex line bundle $L$ such that $c_1(L) \equiv w_2(TM) \pmod{2}$, where $c_1(L)$ is the first Chern class of $L$. The $U(1)$ gauge field $A$ on the D-brane is precisely a connection on such a line bundle...''\end{tcolorbox}

\paragraph{Level 3: Tacit Step Reconstruction.}
\noindent\textit{Success example (Q3, GPT-5.2)}:

\begin{tcolorbox}[modelout]"Consider the $(2+1)$-dimensional $U(1)$ Chern-Simons theory on an oriented $3$-manifold $M$ ... “bosonic” TQFT means it is well-defined on any oriented $3$-manifold without choosing a spin structure. A “fermionic” (spin) TQFT requires a choice of spin structure to define the path integral consistently. The reason $U(1)_k$ is bosonic only for even $k$ is that for odd $k$ the theory necessarily contains a transparent fermion (equivalently, it has nontrivial dependence on spin structure) ... 1) Anyons in $U(1)_k$ and their spins in $U(1)_k$, Wilson lines are labeled by an integer charge class $n \in \mathbb{Z}_k$ ... 2) Existence of a transparent fermion when $k$ is odd ... A key diagnostic for whether a theory is bosonic is whether there exists a local (transparent) fermion: an excitation with trivial mutual braiding with all anyons but with fermionic self-statistics $\theta = -1$. If such a line exists, the theory is naturally a spin-TQFT ... the mutual braiding phase between anyons $n$ and $m$ is $M_{n,m} = \exp\ \left(2\pi i \frac{nm}{k}\right).$ ... Now assume $k$ is odd. Consider the anyon $f = \frac{k+1}{2} \in \mathbb{Z}_k.$ ... (i) $f$ is transparent (trivial mutual braiding)... "\end{tcolorbox}
\noindent\textit{Explanation:} The response presents a coherent reasoning chain linking the
topological obstruction $w_2(M)$, the introduction of a $U(1)$ line bundle,
and the emergence of a Spin$^c$ structure. The conclusion is therefore
derived through an explicit sequence of arguments rather than stated
in isolation, satisfying Level~2.

\subsection{Evaluation Criteria Per Question}
\label{app:eval_criteria}
In this appendix, we present detailed evaluation criteria for Levels~1--3 for all questions.
We do not include criteria for Level~0 and Level~4.
As defined in Section~\ref{sec:scoring criteria}, Level~0 concerns the correctness
of the final statement, which is straightforward to assess.
In contrast, Level~4 (enrichment) is inherently open-ended and does not admit
a unique set of criteria for a given question.
\paragraph{Q1.}
\begin{itemize}
    \item \textbf{L1:} Identifies the relevant conceptual ingredients for the statement, namely spontaneous symmetry breaking, the vacuum expectation value of a scalar order parameter, and the role of the Goldstone mode or Goldstone theorem.

    \item \textbf{L2:} States the key reasoning step that in two-dimensional quantum field theories the would-be order parameter cannot acquire a nonzero vacuum expectation value, so continuous symmetry breaking is excluded.

    \item \textbf{L3:} Explicitly reconstructs the tacit infrared argument, namely that the massless Goldstone fluctuations are infrared divergent in two dimensions, which forces the vacuum expectation value to vanish and thereby prevents spontaneous symmetry breaking.
\end{itemize}

\paragraph{Q2.}
\begin{itemize}
    \item \textbf{L1:} Identifies the relevant conceptual ingredients for the statement, namely two-dimensional conformal field theories, primary operators (or primary states), and two-point correlation functions.

    \item \textbf{L2:} States the key structural claim that spacetime conformal symmetry strongly constrains the form of two-point functions of primary operators in two-dimensional conformal field theories.

    \item \textbf{L3:} Explicitly reconstructs the constraints from individual spacetime symmetries, namely:  
    (a) translation and rotation invariance imply that the correlator depends only on the distance between operator insertions;  
    (b) scale invariance fixes the correlator to a power-law form determined by the conformal dimensions of the operators;  
    (c) special conformal invariance requires the correlator to vanish unless the two operators have identical conformal dimensions.
\end{itemize}

\paragraph{Q3.}
\begin{itemize}
    \item \textbf{L1:} Identifies the relevant conceptual ingredients for the statement, namely Chern--Simons theory and its level, and the distinction between bosonic and spin topological field theories.

    \item \textbf{L2:} States the key structural claim that Chern--Simons theories at odd level contain fermionic anyons (Wilson lines with half-integer spin), or equivalently that the partition function depends on a choice of spin structure.

    \item \textbf{L3:} Explicitly reconstructs the origin of the spin-structure dependence of odd-level Chern--Simons theory in one of the following two ways: (a) by showing that the theory contains a transparent fermionic Wilson line, i.e.\ a half-integer-spin line whose mutual braiding with all bosonic lines is trivial, implying that the theory cannot be consistently defined on general oriented manifolds but requires a choice of spin structure; (b) by extending the $(2+1)$-dimensional spacetime manifold as the boundary of a $(3+1)$-dimensional manifold and analyzing the behavior of the Chern--Simons path integral under large gauge transformations, showing that the action is well-defined only when the spacetime manifold is endowed with a spin structure.
\end{itemize}

\paragraph{Q4.}
\begin{itemize}
    \item \textbf{L1:} Identifies the relevant conceptual ingredients for the statement, namely 1-form symmetries, confinement and deconfinement phases, and the area-law and perimeter-law behaviors of Wilson loops.

    \item \textbf{L2:} States the key structural relation that Wilson lines are charged under 1-form symmetries and that the behavior of their expectation values (area law or perimeter law) diagnoses whether the 1-form symmetry is preserved or spontaneously broken.

    \item \textbf{L3:} Explicitly reconstructs the relation between Wilson loop expectation values and the static potential between external charges, for example by considering a large rectangular loop and showing an area law implies a linear potential and hence confinement.
\end{itemize}

\paragraph{Q5.}
\begin{itemize}
    \item \textbf{L1:} Identifies the relevant conceptual ingredients for the statement, namely the super-Poincaré algebra, supersymmetric vacua, and the scalar potential.

    \item \textbf{L2:} States the key structural claim that supersymmetric vacua are annihilated by the supercharges, which, via the super-Poincaré algebra, implies that the Hamiltonian must vanish and therefore the scalar potential must vanish.

    \item \textbf{L3:} Explicitly reconstructs the argument that a supersymmetric vacuum has vanishing Hamiltonian and that, for a relativistic vacuum preserving spacetime translations, the field configuration must be constant. As a result, the Hamiltonian receives contributions only from the scalar potential. Since the scalar potential is semi-positive definite, it must therefore vanish.
\end{itemize}

\paragraph{Q6.}
\begin{itemize}
    \item \textbf{L1:} Identifies the relevant conceptual ingredients for the statement, namely $(5+1)$-dimensional superconformal field theories and the notion of intrinsically relative quantum field theories. In particular, a relative quantum field theory is defined as a boundary theory for a one-dimension-higher topological quantum field theory.

    \item \textbf{L2:} States the key structural claim that a relative quantum field theory does not possess an ordinary partition function but rather a partition vector, which takes values in the Hilbert space of the one-dimension-higher topological quantum field theory. Moreover, the relativeness of $(5+1)$-dimensional superconformal field theories cannot in general be removed. 

    \item \textbf{L3:} Explicitly reconstructs the intrinsic relativeness of $(5+1)$-dimensional superconformal field theories in one of the following two ways:
    
    (a) the Heisenberg group associated with the three-form flux sector does not, in general, admit a polarization; equivalently, there is in general no maximal mutually local set of flux operators that would define an absolute theory;
    
    (b) the associated $(6+1)$-dimensional topological field theory is a higher-form Chern--Simons theory, which in general does not admit a topological boundary condition that would define a global form of the superconformal field theory.
\end{itemize}

\paragraph{Q7.}
\begin{itemize}
    \item \textbf{L1:} Identifies the relevant conceptual ingredients for the statement, namely a compact boson, Kaluza--Klein momentum modes, winding modes, and T-duality.

    \item \textbf{L2:} States the key structural claim that T-duality acts as a worldsheet transformation (e.g. $X_R \rightarrow -X_R$ together with $R \rightarrow \alpha'/R$) and exchanges momentum and winding modes, without explicitly deriving this exchange.

    \item \textbf{L3:} Explicitly reconstructs how T-duality exchanges momentum and winding by analyzing the zero-mode sector, for instance by writing the left- and right-moving momenta and showing how the transformation maps the momentum and winding contributions into each other.
\end{itemize}

\paragraph{Q8.}
\begin{itemize}
    \item \textbf{L1:} Identifies the relevant conceptual ingredients for the statement, namely the closed-string one-loop amplitude, modular invariance, spin structures (NS and R sectors), and the GSO projection (or summing over spin structures).

    \item \textbf{L2:} States the key structural claim that modular invariance under $SL(2,\mathbb{Z})$ of the worldsheet torus partition function is required for a consistent closed-string one-loop amplitude, and that the GSO projection is imposed to achieve modular invariance.

    \item \textbf{L3:} Explicitly reconstructs how the GSO projection fixes the relative contributions of different spin structures of worldsheet fermions, such that their sum forms a modular invariant partition function.
\end{itemize}

\paragraph{Q9.}
\begin{itemize}
    \item \textbf{L1:} Identifies the relevant conceptual ingredients for the statement, namely Chan--Paton factors, one-dimensional Majorana fermions on the boundary, Clifford algebras, and the Lie algebras $\mathfrak{u}(n)$, $\mathfrak{so}(n)$, and $\mathfrak{sp}(n)$.

    \item \textbf{L2:} States the key structural claim that one-dimensional boundary Majorana fermions, through their Clifford algebra, can be used to construct generators of $\mathfrak{u}(n)$, $\mathfrak{so}(n)$, and $\mathfrak{sp}(n)$.

    \item \textbf{L3:} Explicitly reconstructs how bilinear currents of boundary Majorana fermions generate Lie algebras, by showing how different reality conditions (real, complex, symplectic) lead to $\mathfrak{so}(n)$, $\mathfrak{u}(n)$, and $\mathfrak{sp}(n)$ via deriving their commutation relations.
\end{itemize}

\paragraph{Q10.}
\begin{itemize}
    \item \textbf{L1:} Identifies the relevant conceptual ingredients for the statement, namely D-brane worldvolumes, spin structures, Spin$^c$ structures, and $U(1)$ gauge fields.

    \item \textbf{L2:} States the key structural claim that fermions on the D-brane worldvolume need not require a spin structure; instead, in the presence of a $U(1)$ gauge field, the obstruction to defining spinors can be compensated by the associated line bundle, so that the relevant structure is a Spin$^c$ structure, without explicitly reconstructing the obstruction or its cancellation.

    \item \textbf{L3:} Explicitly reconstructs the mechanism that the second Stiefel--Whitney class $w_2$ obstructs the definition of spinors, leading to a sign ambiguity in the fermion path integral, and shows how coupling to a $U(1)$ gauge field whose first Chern class $c_1$ cancels this obstruction (e.g.\ $w_2 \equiv c_1 \!\!\mod 2$), thereby defining a Spin$^c$ structure.
\end{itemize}

\paragraph{Q11.}
\begin{itemize}
    \item \textbf{L1:} Identifies the relevant conceptual ingredients for the statement, namely 't Hooft anomalies, chiral (ABJ) anomalies, and renormalization group flows.

    \item \textbf{L2:} States the key structural claim that the apparent contradiction is resolved by distinguishing two different types of anomalies: chiral (ABJ) anomalies, which break the symmetry at the quantum level, and 't Hooft anomalies, which do not break the symmetry but instead characterize it.

    \item \textbf{L3:} Explicitly explains that a symmetry with a chiral (ABJ) anomaly is not a genuine global symmetry at the quantum level, as its current is not conserved, whereas a symmetry with a 't Hooft anomaly remains a well-defined global symmetry with a conserved current but cannot be consistently gauged. Furthermore, the 't Hooft anomaly is an intrinsic property of the symmetry and is invariant under renormalization group flow, thereby enabling anomaly matching between the ultraviolet and infrared.
\end{itemize}

\paragraph{Q12.}
\begin{itemize}
    \item \textbf{L1:} Identifies the relevant conceptual ingredients for the statement, namely spontaneous symmetry breaking, degenerate classical vacua, quantum tunneling, and Hamiltonian eigenstates.

    \item \textbf{L2:} States the key structural claim that in finite volume quantum tunneling mixes different classical vacua, leading to a unique symmetric ground state, whereas in infinite volume the tunneling amplitude is suppressed, resulting in superselection sectors and allowing spontaneous symmetry breaking.

    \item \textbf{L3:} Explicitly reconstructs the mechanism that when the Hamiltonian commutes with the symmetry operator, its exact ground state must be symmetric and is given by a superposition of classically degenerate vacua in finite volume. It then shows that the tunneling amplitude between these vacua is exponentially suppressed with the spatial volume (e.g.\ $\sim e^{-c\,V}$), so that in the infinite-volume limit the mixing vanishes, leading to degenerate ground states and superselection sectors, thereby permitting spontaneous symmetry breaking.
\end{itemize}

\section{Evaluation}

\subsection{Performance Across Domain Landscape}
\label{app:domain}
Table~\ref{tab:domain_landscape} summarizes performance across broad
theoretical physics domains after averaging scores over the five evaluation
levels (L0--L4). The results can be interpreted along two complementary axes:
variation across domains and variation across models.

\begin{table}[ht!]
\centering
\small
\setlength{\tabcolsep}{4pt}
\renewcommand{\arraystretch}{1.1}
\begin{tabular}{lccccc}
\toprule
Model 
& FT foundations 
& Symmetry 
& Conformal
& SUSY
& Strings\\ 
\midrule

Gemini-2.5-flash
& \cellcolor{yellow!25}2.500
& \cellcolor{yellow!35}2.800
& \cellcolor{yellow!35}2.750
& \cellcolor{yellow!35}2.750
& \cellcolor{yellow!25}2.667 \\

Gemini-3.1-pro-preview
& \cellcolor{green!25}4.750
& \cellcolor{green!25}4.400
& \cellcolor{green!25}4.250
& \cellcolor{green!25}4.500
& \cellcolor{green!25}4.333 \\

GPT-5.2
& \cellcolor{green!25}4.250
& \cellcolor{green!15}3.600
& \cellcolor{green!15}3.750
& \cellcolor{green!15}3.750
& \cellcolor{green!15}3.667 \\

GPT-4.1
& \cellcolor{yellow!35}3.250
& \cellcolor{yellow!35}3.000
& \cellcolor{yellow!35}3.250
& \cellcolor{yellow!35}3.250
& \cellcolor{yellow!35}3.333 \\

Deepseek-V3.2 (non-thinking)
& \cellcolor{yellow!35}3.000
& \cellcolor{yellow!25}2.600
& \cellcolor{yellow!35}2.750
& \cellcolor{yellow!35}2.750
& \cellcolor{yellow!35}3.000 \\

Deepseek-V3.2 (thinking)
& \cellcolor{yellow!35}3.000
& \cellcolor{yellow!35}2.800
& \cellcolor{yellow!35}3.000
& \cellcolor{yellow!35}3.000
& \cellcolor{yellow!35}3.000 \\

Kimi-K2-thinking
& \cellcolor{yellow!25}2.667
& \cellcolor{yellow!25}2.000
& \cellcolor{yellow!35}3.250
& \cellcolor{yellow!35}2.750
& \cellcolor{yellow!35}3.000 \\

Qwen3.5-397b
& \cellcolor{yellow!35}2.833
& \cellcolor{yellow!25}2.600
& \cellcolor{green!15}3.750
& \cellcolor{yellow!35}3.000
& \cellcolor{green!15}3.667 \\

Minimax-m2.7
& \cellcolor{yellow!25}2.000
& \cellcolor{yellow!25}2.000
& \cellcolor{yellow!25}2.500
& \cellcolor{yellow!25}2.250
& \cellcolor{yellow!25}2.667 \\

Nemotron-3-super
& \cellcolor{red!20}1.667
& \cellcolor{yellow!25}2.000
& \cellcolor{yellow!25}2.500
& \cellcolor{yellow!25}2.500
& \cellcolor{yellow!35}3.000 \\

\bottomrule
\end{tabular}
\caption{
Average model performance by broad research area. For each question, scores are first summed across the five evaluation levels (L0--L4), and then averaged over all questions belonging to that research area. Because the research areas overlap, some questions contribute to multiple columns. Colors indicate performance from low (red) to high (green) on a uniform scale (score\,/\,5).
}
\label{tab:domain_landscape}
\end{table}

\noindent\textbf{Variation across domains.}
Performance varies across research areas, but the variation is moderate
compared to the differences observed across evaluation levels.
Problems in field theory foundations and symmetry and topological structures
tend to be more challenging, with several models achieving scores in the range
of $2.0$--$3.0$.
These domains often involve global constraints and structural distinctions
that are not fully captured by local reasoning alone.

In contrast, conformal and lower-dimensional QFT problems consistently yield
higher performance across models, with many systems achieving scores above $3.5$.
This suggests that reasoning within well-established and internally consistent
formal frameworks is more robustly handled.
Problems involving string dualities and D-brane physics also show relatively
strong performance, typically clustering around $3.0$--$3.7$,
indicating that certain recurring structural patterns in this domain are
effectively learned.

SUSY and higher-dimensional QFT problems occupy an intermediate regime,
with performance generally stable across models but not reaching the highest
scores observed in conformal or string-theoretic tasks.

\noindent\textbf{Variation across models.}
Across all domains, leading models such as G3.1-pro-preview and GPT-5.2
demonstrate consistently strong performance, maintaining scores above $3.5$
in most domains and reaching values above $4.0$ in several cases.
In contrast, smaller or open-weight models exhibit lower overall performance
and greater variability.
For example, models such as Minimax-2.7 and Nemo-3 achieve scores near $2.0$
in more challenging domains while improving to around $2.5$--$3.0$
in comparatively easier ones.

Intermediate models, including GPT-4.1 and Deepseek-3.2 variants, display relatively
stable but moderate performance across domains, typically clustering around
$3.0$.
Notably, domain sensitivity is present but limited:
most models vary by less than one point across different research areas.

Taken together, these results indicate that while domain structure does
influence performance, it is not the primary source of difficulty.
Instead, the relatively narrow spread across domains suggests that
performance is governed more strongly by the type of reasoning required,
a point we analyze further in the next subsection.

\subsection{Reasoning Geometry}
\label{app:reasoning geometry}
\noindent\textbf{Local derivation tasks.}
In the mechanism-driven and single-structure quadrant,
performance is near-saturated at Levels~0--2 across all models,
indicating that explicit derivations within a stable conceptual frame
are reliably handled.
Separation begins at Level~3, where models must reconstruct tacit
intermediate steps, but degradation remains moderate (Table~\ref{tab:local_derivation}).
This suggests that models are effective at sequentially expanding
well-structured reasoning chains when the underlying representation
is fixed.


\begin{table}[ht!]
\centering
\begin{tabular}{lccccc}
\toprule
Model & L0 & L1 & L2 & L3 & L4 \\
\midrule

Gemini-2.5-flash 
& 1.000 & 1.000 & 1.000 & 0.250 & 0.000 \\

Gemini-3.1-pro-preview 
& 1.000 & 1.000 & 1.000 & 1.000 & 0.500 \\

GPT-5.2 
& 1.000 & 1.000 & 1.000 & 0.750 & 0.000 \\

GPT-4.1 
& 1.000 & 1.000 & 1.000 & 0.500 & 0.000 \\

Deepseek-V3.2 (non-thinking) 
& 1.000 & 1.000 & 0.750 & 0.250 & 0.000 \\

Deepseek-V3.2 (thinking) 
& 1.000 & 1.000 & 0.750 & 0.500 & 0.250 \\

Kimi-K2-thinking 
& 1.000 & 1.000 & 1.000 & 0.750 & 0.000 \\

Qwen3.5-397b 
& 1.000 & 1.000 & 1.000 & 0.500 & 0.250 \\

Minimax-m2.7 
& 1.000 & 1.000 & 0.500 & 0.250 & 0.000 \\

Nemotron-3-super 
& 1.000 & 0.750 & 0.500 & 0.250 & 0.000 \\

\bottomrule
\end{tabular}
\caption{Average performance on Local Derivation tasks across evaluation levels (L0--L4).}
\label{tab:local_derivation}
\end{table}

\medskip

\noindent\textbf{Integration tasks.} 
When remaining mechanism-driven but requiring cross-domain synthesis, performance begins to differentiate.
While Levels~0--2 remain relatively strong,
substantial variation appears at Level~3 (Table~\ref{tab:integration}).
This indicates that the primary difficulty lies not in executing
individual derivations, but in combining distinct reasoning threads
into a coherent argument.
Nevertheless, performance remains significantly higher than in
consistency-driven regimes, suggesting that the conceptual frame
remains largely stable .


\begin{table}[ht!]
\centering
\begin{tabular}{lccccc}
\toprule
Model & L0 & L1 & L2 & L3 & L4 \\
\midrule

Gemini-2.5-flash 
& 1.000 & 0.667 & 0.667 & 0.000 & 0.000 \\

Gemini-3.1-pro-preview 
& 1.000 & 1.000 & 1.000 & 1.000 & 0.667 \\

GPT-5.2 
& 1.000 & 1.000 & 1.000 & 0.667 & 0.000 \\

GPT-4.1 
& 1.000 & 1.000 & 1.000 & 0.333 & 0.000 \\

Deepseek-V3.2 (non-thinking) 
& 1.000 & 1.000 & 0.667 & 0.333 & 0.000 \\

Deepseek-V3.2 (thinking) 
& 1.000 & 0.667 & 0.667 & 0.333 & 0.000 \\

Kimi-K2-thinking 
& 1.000 & 0.667 & 0.667 & 0.333 & 0.000 \\

Qwen3.5-397b 
& 1.000 & 1.000 & 1.000 & 0.667 & 0.333 \\

Minimax-m2.7 
& 1.000 & 0.667 & 0.667 & 0.000 & 0.000 \\

Nemotron-3-super 
& 1.000 & 0.667 & 0.667 & 0.000 & 0.000 \\
\bottomrule
\end{tabular}
\caption{Average performance on Integration tasks across evaluation levels (L0--L4).}
\label{tab:integration}
\end{table}

\medskip

\noindent\textbf{Constraint-based reasoning tasks.}
In the consistency-driven but single-structure quadrant,
performance degradation appears earlier.
Even at Level~2 (Table~\ref{tab:constraint}), several models show noticeable decline,
indicating that enforcing global consistency conditions introduces
additional reasoning complexity.
At Level~3, models diverge sharply:
while stronger systems can propagate constraints through the argument,
weaker models often fail to complete the tacit reconstruction.
This suggests that recognizing a constraint is insufficient;
the difficulty lies in integrating it consistently into the reasoning process .


\begin{table}[ht!]
\centering
\begin{tabular}{lccccc}
\toprule
Model & L0 & L1 & L2 & L3 & L4 \\
\midrule

Gemini-2.5-flash 
& 1.000 & 0.667 & 0.667 & 0.000 & 0.000 \\

Gemini-3.1-pro-preview 
& 1.000 & 1.000 & 1.000 & 0.667 & 0.333 \\

GPT-5.2 
& 1.000 & 1.000 & 0.667 & 0.667 & 0.000 \\

GPT-4.1 
& 1.000 & 1.000 & 0.667 & 0.333 & 0.000 \\

Deepseek-V3.2 (non-thinking) 
& 1.000 & 1.000 & 0.333 & 0.333 & 0.000 \\

Deepseek-V3.2 (thinking) 
& 1.000 & 1.000 & 0.667 & 0.333 & 0.000 \\

Kimi-K2-thinking 
& 1.000 & 1.000 & 0.667 & 0.333 & 0.000 \\

Qwen3.5-397b 
& 1.000 & 1.000 & 1.000 & 0.667 & 0.000 \\

Minimax-m2.7 
& 1.000 & 0.667 & 0.667 & 0.333 & 0.000 \\

Nemotron-3-super 
& 1.000 & 0.667 & 0.667 & 0.333 & 0.000 \\

\bottomrule
\end{tabular}
\caption{Average performance on Constraint-based tasks across evaluation levels (L0--L4).}
\label{tab:constraint}
\end{table}

\medskip

\noindent\textbf{Conceptual hinge tasks.}
The most substantial degradation occurs in the quadrant combining
consistency-driven inference with multi-structure organization (Table~\ref{tab:hinge}).
Performance drops already at Level~2 for many models,
and collapses almost entirely at Level~3.
These tasks require identifying a latent structural distinction
that resolves an apparent conceptual tension before any derivation proceeds.
Unlike other regimes, failure occurs not in executing reasoning steps,
but in initiating the correct representational frame.


\begin{table}[ht!]
\centering
\begin{tabular}{lccccc}
\toprule
Model & L0 & L1 & L2 & L3 & L4 \\
\midrule

Gemini-2.5-flash 
& 1.000 & 1.000 & 0.500 & 0.000 & 0.000 \\

Gemini-3.1-pro-preview 
& 1.000 & 1.000 & 1.000 & 1.000 & 0.500 \\

GPT-5.2 
& 1.000 & 1.000 & 1.000 & 0.500 & 0.000 \\

GPT-4.1 
& 1.000 & 1.000 & 1.000 & 0.000 & 0.000 \\

Deepseek-V3.2 (non-thinking) 
& 1.000 & 1.000 & 0.500 & 0.000 & 0.000 \\

Deepseek-V3.2 (thinking) 
& 1.000 & 1.000 & 0.500 & 0.000 & 0.000 \\

Kimi-K2-thinking 
& 1.000 & 1.000 & 0.000 & 0.000 & 0.000 \\

Qwen3.5-397b 
& 1.000 & 0.500 & 0.500 & 0.000 & 0.000 \\

Minimax-m2.7 
& 1.000 & 0.500 & 0.000 & 0.000 & 0.000 \\

Nemotron-3-super 
& 1.000 & 1.000 & 0.000 & 0.000 & 0.000 \\

\bottomrule
\end{tabular}
\caption{Average performance on Conceptual Hinge tasks across evaluation levels (L0--L4).}
\label{tab:hinge}
\end{table}
\end{document}